\documentclass[conference]{IEEEtran}
\IEEEoverridecommandlockouts

\usepackage{cite}
\usepackage{amsmath,amssymb,amsfonts}
\usepackage{algorithm}
\usepackage{graphicx}
\usepackage{textcomp}
\usepackage{xcolor}
\usepackage[T1]{fontenc}
\usepackage[utf8]{inputenc}
\usepackage{tabularx,booktabs,ragged2e}
\newcolumntype{L}{>{\RaggedRight\arraybackslash}X}
\newcolumntype{R}{>{\RaggedLeft\arraybackslash}X}
\usepackage{multirow}
\usepackage{algpseudocode}

\def\BibTeX{{\rm B\kern-.05em{\sc i\kern-.025em b}\kern-.08em
    T\kern-.1667em\lower.7ex\hbox{E}\kern-.125emX}}
\begin{document}

\title{
    Unsupervised Anomaly Detection in NSL-KDD Using $\beta$-VAE: A Latent Space and Reconstruction Error Approach
    \thanks{
        The work presented in this article was carried out as part of collaboration between CReSTIC Labs and Seckiot funded under the ANRT (Association Nationale de la Recherche et de la Technologie), the national association for Research and Technology.
    }
}

\author{%
    \IEEEauthorblockN{
        Dylan Baptiste\IEEEauthorrefmark{1}\IEEEauthorrefmark{2},
        Ramla Saddem\IEEEauthorrefmark{1},
        Alexandre Philippot\IEEEauthorrefmark{1},
        François Foyer\IEEEauthorrefmark{2}
    }
    \IEEEauthorblockA{
        \IEEEauthorrefmark{1}Université de Reims Champagne-Ardenne, CRESTIC, Reims, France\\
        Email: \{dylan.baptiste, ramla.saddem, alexandre.philippot\}@univ-reims.fr
    }
    \IEEEauthorblockA{
        \IEEEauthorrefmark{2}Seckiot, Paris, France\\
        Email: francois.foyer@seckiot.fr
    }
}

\maketitle

\begin{abstract}
As Operational Technology increasingly integrates with Information Technology, the need for Intrusion Detection Systems becomes more important. This paper explores an unsupervised approach to anomaly detection in network traffic using $\beta$-Variational Autoencoders on the NSL-KDD dataset. We investigate two methods: leveraging the latent space structure by measuring distances from test samples to the training data projections, and using the reconstruction error as a conventional anomaly detection metric. By comparing these approaches, we provide insights into their respective advantages and limitations in an unsupervised setting. Experimental results highlight the effectiveness of latent space exploitation for classification tasks.
\end{abstract}


\section{Introduction}

The increasing integration of Operational Technology (OT) with Information Technology (IT) systems has led to a growing need for intrusion detection systems (IDS) in industrial environments. Anomaly detection is a crucial component of IDS, as it enables the identification of malicious activities that deviate from normal behavior. In this context, the NSL-KDD dataset \cite{tavallaee2009detailed} is a widely used benchmark for evaluating the performance of anomaly detection algorithms.

In this paper, we explore an unsupervised approach to anomaly detection in network traffic using $\beta$-Variational Autoencoders ($\beta$-VAE) \cite{higgins2017beta}. $\beta$-VAEs are a deep learning model that can learn a low-dimensional representation of the input data, known as the latent space. By leveraging the latent space structure, we aim to detect anomalies in network traffic without the need for labeled data.

We investigate two complementary methods for anomaly detection using $\beta$-VAEs: measuring the distances from test samples to the projections of the training data in the latent space, and using the reconstruction error as a conventional anomaly detection metric. By comparing these approaches, we provide insights into their respective advantages and limitations in an unsupervised setting.

The remainder of this paper is organized as follows: Section~\ref{sec:definitions} provides definitions and formalizations of the key concepts in this study, such as the $\beta$-VAE model and the NSL-KDD dataset. Section~\ref{sec:related_work} presents related work in the field of anomaly detection and the use of the latent space in the autoencoder framework. Section~\ref{sec:methodology} describes the methodology used in this study, including the VAE architecture and the anomaly detection methods. Section~\ref{sec:results} presents the experimental results and discusses the performance of the proposed methods. Finally, Section~\ref{sec:conclusion} concludes the paper and outlines directions for future work.

\section{Definitions} \label{sec:definitions}

\begin{table*}[ht]
    \caption{Categories of attacks in NSL-KDD}
    \label{tab:attacks}
    \renewcommand{\tabcolsep}{3pt} 
    \begin{tabularx}{\textwidth}{LLLL}
        \toprule
        \textbf{DoS} & \textbf{Probe} & \textbf{U2R} & \textbf{R2L} \\
        \midrule
        neptune, smurf, back, teardrop, pod, land, apache2, mailbomb, processtable, udpstorm, worm & ipsweep, nmap, portsweep, satan, mscan, saint & buffer\_overflow, loadmodule, perl, rootkit, httptunnel, ps, sqlattack, xterm & ftp\_write, guess\_passwd, imap, multihop, phf, spy, warezclient, warezmaster, snmpgetattack, snmpguess, xlock, xsnoop \\
        \bottomrule
    \end{tabularx}
\end{table*}

\subsection{$\beta$-VAE Model} \label{sec:beta_vae_model}

    The $\beta$-VAE architecture is composed of an \textit{encoder} $q$ that maps the input data $x$ to the latent space $z$ and a \textit{decoder} $p$ that reconstructs the initial data from the latent representation.

    The original training objective of a VAE is to maximize the evidence lower bound (ELBO), which for a $\beta$-VAE becomes
    \begin{equation}
        \mathbb{E}_{q_{\phi}(z|x)}[\log p_{\theta}(x|z)] - \beta D_{KL}(q_{\phi}(z|x) \Vert p(z)) \label{eq:beta_vae_elbo}
    \end{equation}

    where $\theta$ and $\phi$ are the parameters of the decoder and encoder networks, respectively, and $z$ is the latent variable; $q_{\phi}(z|x)$ is the approximate posterior distribution of $z$ given input $x$, $p_{\theta}(x|z)$ is the data likelihood given the latent variable, and $D_{KL}$ is the Kullback–Leibler divergence \cite{kullback1951information} between the approximate posterior and the prior distribution $p(z)$.
    The $\beta$ term controls the balance between latent space regularization and reconstruction fidelity. Higher values of $\beta$ enforce greater disentanglement but may compromise reconstruction accuracy.
    The model is trained by minimizing the loss function~\eqref{eq:beta_vae_loss} with a stochastic gradient descent algorithm.

    In practice, the model is trained by minimizing the negative ELBO, which is written as
    \begin{equation}
        - \,\mathbb{E}_{q_{\phi}(z|x)}[\log p_{\theta}(x|z)] + \beta D_{KL}(q_{\phi}(z|x) \Vert p(z)) \label{eq:beta_vae_loss} 
    \end{equation}

    A reparameterization trick \cite{kingma2022autoencoding} is applied to sample $z$ from the latent distribution $q_{\phi}(z|x)$, allowing gradients to be backpropagated through the stochastic sampling process.
    
    In the rest of this study the reconstruction error will be denoted as $\mathcal{L}_{\text{rec}} = - \,\mathbb{E}_{q_{\phi}(z|x)}[\log p_{\theta}(x|z)]$ and the KL divergence as $\mathcal{L}_{\text{KL}} = D_{KL}(q_{\phi}(z|x) \Vert \mathcal{N}(0, I))$, where $\mathcal{N}(0, I)$ is the standard normal distribution.

    \subsection{NSL-KDD Dataset} \label{sec:nsl_kdd}

    The NSL-KDD dataset is a benchmark dataset for evaluating the performance of intrusion detection systems. It is a modified version of the KDD Cup 1999 dataset, which contains network traffic data. The NSL-KDD dataset consists of 41 features, including 34 continuous and 7 categorical features. The dataset contains five classes of network traffic: normal, denial of service (DoS), probe, user-to-root (U2R), and remote-to-local (R2L). Table~\ref{tab:attacks} shows the categories of attacks in the NSL-KDD dataset \cite{tavallaee2009detailed}.

\section{Related Work} \label{sec:related_work}

    Anomaly detection in unsupervised settings has been the subject of numerous approaches based on autoencoders (AE) and their variants, with some recent studies focusing on leveraging the latent space.
    Previous works on unsupervised anomaly detection have explored various methods to handle high-dimensional or highly nonlinear data.

    Already in 2007, \cite{HeFault2007} proposed a fault detection method for industrial processes based on the k-nearest neighbors (k-NN) rule, using only data from normal operation. This approach addresses the absence of anomalous training data by modeling the distribution of distances between normal samples and their nearest neighbors. Anomalies are then identified as samples whose distance exceeds a threshold derived from this distribution.
    In this work, the authors clearly explain the principle that will be revisited later in works exploiting approaches generating richer spaces, such as the latent spaces of AEs or their variants.

    Hybrid models combining AEs or their variants with neighborhood techniques have been developed to enhance anomaly detection on high-dimensional data \cite{SongHybrid2017}, \cite{guoAnomaly2018}, and \cite{angiullilatent2023}. These methods highlighted the advantage of nonlinear representation while leveraging distance measures in the latent space.

    Other works focused on industrial applications and monitoring systems have also highlighted the interest of this approach, demonstrating that combining reconstruction error with latent space distance analysis can yield competitive or even superior performance compared to traditional anomaly detection methods \cite{zhangAutomated2018} and \cite{corizzoAnomaly2019}.

    Subsequently, works introduced the use of models based on Variational Autoencoders (VAE), and more specifically $\beta$-VAE, to achieve a more structured and interpretable organization of the latent space. These studies suggest that considering the distribution of latent variables can contribute to finer anomaly detection, whether through reconstruction error measures or distances in the latent space \cite{ramakrishnaEfficient2021}. In a similar context, \cite{pitsiorlasTrustworthy2024} aimed at estimating a confidence measure through the exploitation of projections in the latent space and Mahalanobis distance has been presented to enhance intrusion detection on datasets like NSL-KDD.

    Other contributions have sought to constrain the latent space to promote the emergence of clusters with similar behaviors. For example, \cite{astridConstricting2024} aims to limit the reconstruction capacity of AEs during training using an additional constraint that acts as a regularization on the latent space.
    In \cite{kamnitsasSemi2018}, compact clustering methods in the latent space were developed in a semi-supervised framework, allowing for the grouping of projections of samples with the same label, attracting unlabeled projections in the space and thus better identifying deviations. These techniques illustrate the interest of latent structure for separating normal data from anomalies.

    The work proposed in this article aligns with the previously mentioned studies. We leverage the structure of the latent space of a $\beta$-VAE for anomaly detection in network traffic, using the NSL-KDD dataset. We formalize and compare the two methods (reconstruction error and distance in the latent space) for anomaly detection, highlighting their respective advantages and disadvantages.
    We show that exploiting the latent space yields results comparable to those of reconstruction error, while providing better interpretability of the results.
    We also observe the impacts of certain parameters on the performance of both methods, particularly $\beta$ and $k$, the number of neighbors considered for calculating the distance in the latent space.

\section{Methodology} \label{sec:methodology}

    In this study, we focus on the binary classification task of detecting normal traffic and anomalies in an unsupervised setting; in the analysis, we will also present results per attack classes and categories.
    
    \subsection{Data Preprocessing}
    We exploit an unsupervised learning approach, using only the labeled \textit{normal} data from the training dataset.
    We have therefore re-divided NSL-KDD presented in section~\ref{sec:nsl_kdd} into 3 parts:
    \begin{itemize}
        \item The anomalous dataset, which includes all attack data from the training and test datasets. This dataset is named $X_{attack}$.
        \item The training dataset, which includes only the labeled \textit{normal} data from the training dataset. This dataset is named $X_{train}$.
        \item The test dataset, which includes only the labeled \textit{normal} data from the test dataset. This dataset is named $X_{test}$.
    \end{itemize}

    The NSL-KDD dataset includes a mix of categorical, boolean, and continuous features.
    To prepare the data for training the $\beta$-VAE model, we first applied one-hot encoding to the $3$ categorical features (\textit{protocol\_type}, \textit{service}, and \textit{flag}), converting them into binary vectors.
    The $4$ boolean features (\textit{land}, \textit{logged\_in}, \textit{is\_guest\_login}, and \textit{is\_host\_login}) were encoded as binary values (0 or 1).
    The remaining $33$ features are continuous and were standardized using the mean and standard deviation computed from the training set $X_{train}$, ensuring all features operate on a comparable scale, which is essential for stable model training.
    
    No feature selection was performed; we retained all features except for the \textit{difficulty} attribute, which is not relevant to our study. Class labels were excluded during training to maintain an unsupervised learning setting.
    
    \subsection{Model Architecture} \label{sec:model_architecture}

    We use a $\beta$-VAE architecture as described in section~\ref{sec:beta_vae_model}.
    The encoder and decoder networks each comprise three fully connected hidden layers: $64$, $32$, and $16$ neurons for the encoder, and $16$, $32$, and $64$ neurons for the decoder.
    The encoder outputs the mean and log variance through two separate fully connected layers with $8$ neurons each, which represent the parameters of the Gaussian distribution in this $8$-dimensional latent space.
    The decoder takes the sampled latent variable and reconstructs the input data.
    For the stochastic gradient descent algorithm, we use the Adam optimizer \cite{kingmaAdam2017} with a learning rate of $0.001$ and a batch size of $2048$.

    During reconstruction, the total reconstruction loss is computed as a linear combination of three loss functions, each tailored to the nature of the feature:
    \begin{itemize}
        \item \textbf{Categorical features:} softmax activation with categorical cross-entropy loss
        \begin{equation} \label{eq:cat_loss}
        \mathcal{L}_{\text{cat}} = -\frac{1}{n} \sum_{i=1}^{n} \sum_{j=1}^{m} x_{ij} \log(\hat{x}_{ij})
        \end{equation}
        where $n$ is the number of samples, $m$ is the number of categories for a given feature, $x_{ij}$ is a binary indicator (0 or 1) that the $i^{th}$ sample belongs to category $j$, and $\hat{x}_{ij}$ is the predicted probability for category $j$ in sample $i$.
    
        \item \textbf{Boolean features:} sigmoid activation with binary cross-entropy loss
        \begin{equation} \label{eq:bool_loss}
        \mathcal{L}_{\text{bool}} = -\frac{1}{n} \sum_{i=1}^{n} \left[ x_i \log(\hat{x}_i) + (1 - x_i) \log(1 - \hat{x}_i) \right]
        \end{equation}
        where $x_i$ is the true binary value and $\hat{x}_i$ is the predicted probability for the $i^{th}$ sample.
    
        \item \textbf{Continuous features:} linear activation with mean squared error (MSE)
        \begin{equation} \label{eq:cont_loss}
        \mathcal{L}_{\text{cont}} = \frac{1}{n} \sum_{i=1}^{n} (x_i - \hat{x}_i)^2
        \end{equation}
        where $x_i$ and $\hat{x}_i$ are the true and reconstructed continuous values for the $i^{th}$ sample.
    \end{itemize}
    
    The total reconstruction loss $\mathcal{L}_{\text{rec}}$ is defined as a linear combination of the three components
    \begin{equation}
        \mathcal{L}_{\text{rec}} = \mathcal{L}_{\text{cat}} + \mathcal{L}_{\text{bool}} + \mathcal{L}_{\text{cont}} \label{eq:reconloss}
    \end{equation}

    The $\beta$-VAE loss $\mathcal{L}$ is then the combination of $\mathcal{L}_{\text{rec}}$ and the KL divergence term, as defined in Equation~\eqref{eq:beta_vae_loss} and weighted by the $\beta$ parameter
    \begin{equation} \label{eq:beta_vae_loss_final}
        \mathcal{L} = \mathcal{L}_{\text{rec}} + \beta \, \mathcal{L}_{\text{KL}}
    \end{equation}


\section{$\beta$-VAE exploitation for classification} \label{sec:beta_vae_exploitation}

Anomaly detection can be approached in two distinct ways within the framework of our $\beta$-VAE model: through reconstruction error or by analyzing the latent space. Each of these methods allows classifying data as anomalies or normal data, but based on different criteria.
In both cases, we evaluate performance using false positive rate (FPR) and true positive rate (TPR) for different thresholds, with the Area Under the Receiver Operating Characteristic curve (AUROC) as the performance metric.

\subsection{Anomaly detection based on reconstruction error}

    The first anomaly detection approach relies on reconstruction error, a classic method in unsupervised learning.
    After training the $\beta$-VAE model, each data point from the set $X_{attack} \cup X_{test}$ is projected into the latent space using the encoder $z \sim q_\phi(z|x) $, and then reconstructed by the decoder $\hat{x} = p_{\theta}(x|z)$.
    The goal is to quantify the difference between the original data $x$ and its reconstruction $\hat{x}$ from the latent space.
    This difference is measured by the reconstruction error $\mathcal{L}_{\text{rec}}$ presented in Section~\ref{sec:model_architecture}.

    Once the reconstruction error is calculated, a threshold is set to distinguish normal data from anomalous data.
    Data points for which the error exceeds this threshold are considered anomalous, while those with an error below the threshold are classified as normal.
    This detection approach is named $\mathcal{L}_{rec}$-classification in the rest of this work.

    The Algorithm~\ref{alg:recon_classif} implements $\mathcal{L}_{rec}$-classification.

    \begin{algorithm}[h]
        \caption{$\mathcal{L}_{rec}$-classification} \label{alg:recon_classif}
        \begin{algorithmic}[1]
            \Require $x$ a sample to classify, ($q_\phi$, $p_\theta$) : a trained $\beta$-VAE, $\tau$ : the threshold
            \Ensure $y$ : classification label : normal or anomaly
            \State $z \sim q_\phi(z|x)$ \Comment{Encoder}
            \State $\hat{x} \gets p_\theta(z)$ \Comment{Decoder}
            \If{$\mathcal{L}_{\text{rec}}(x, \hat{x}) > \tau$}
            \State $y \gets$ anomaly
            \Else
            \State $y \gets$ normal
            \EndIf
            \State \Return $y$
        \end{algorithmic}
    \end{algorithm}
    
    \subsection{Anomaly detection based on latent space}

    The second approach involves leveraging the latent space of the $\beta$-VAE model to detect anomalies.
    The idea is to project the normal data from the training set $X_{train}$ into the latent space using the encoder $q_\phi$ of the $\beta$-VAE model.
    
    This approach will be referred to as $\mathcal{Z}_k$-classification, where $k$ is an integer representing the number of neighbors to consider for calculating the average Euclidean distance.
    We denote $\mathcal{Z}^{X}_{k}(x)$ as the average Euclidean distance between $z$ (the projection of $x$) and the $k$ nearest neighbors of the projections of $X$. This average is calculated using the formula~\eqref{eq:z_k_x}.

    \begin{equation}
        \mathcal{Z}^{X}_{k}(x) = \frac{1}{k} \sum_{j=1}^{k} \| z - z'_{(j)} \|_2\label{eq:z_k_x}
    \end{equation}
    
    with $z \sim q_\phi(z|x)$ and $z'_{(j)}$ the $j$-th nearest neighbor of $z$ in the set of projections of $X$.
    
    Similarly to the reconstruction error-based method, if an average distance exceeds a threshold, the data point will be considered anomalous; otherwise, it will be classified as normal.
    
    The Algorithm~\ref{alg:zk_classif} implements the $\mathcal{Z}_{k}$-classification.

    \begin{algorithm}[h]
        \caption{$\mathcal{Z}_{k}$-classification} \label{alg:zk_classif}
        \begin{algorithmic}[1]
            \Require $x$ a sample to classify, ($q_\phi$, $p_\theta$) : a trained $\beta$-VAE, $X_{train}$ : the training dataset, $k$ : the number of neighbors, $\tau$ : the threshold
            \Ensure $y$ : classification label : normal or anomaly
            \State $Z_{train} \gets \{z_i \sim q_\phi(z|x_i), \forall x_i \in X_{train} \}$
            \State $z \sim q_\phi(z|x)$
            \State Find the $k$ nearest neighbors $z'_{(1)}, \dots, z'_{(k)}$ of $z$ in $Z_{train}$
            \If{$\mathcal{Z}^{X_{train}}_{k}(x) > \tau$}
                \State $y \gets$ anomaly
            \Else
                \State $y \gets$ normal
            \EndIf
            \State \Return $y$
        \end{algorithmic}
    \end{algorithm}
    

\section{Experimental results} \label{sec:results}

First, we present the performance of the two methods, $\mathcal{L}_{rec}$-classification and $\mathcal{Z}_{k}$-classification, on the binary classification of normal versus anomalous traffic.
As stated in Section~\ref{sec:beta_vae_exploitation}, we evaluate performance using AUROC.
To assess stability, we ran both methods over four runs with different seeds.
The AUROC is computed for each run and then averaged over the four runs.

\begin{figure}[h] \centering
    \includegraphics[width=0.9\columnwidth]{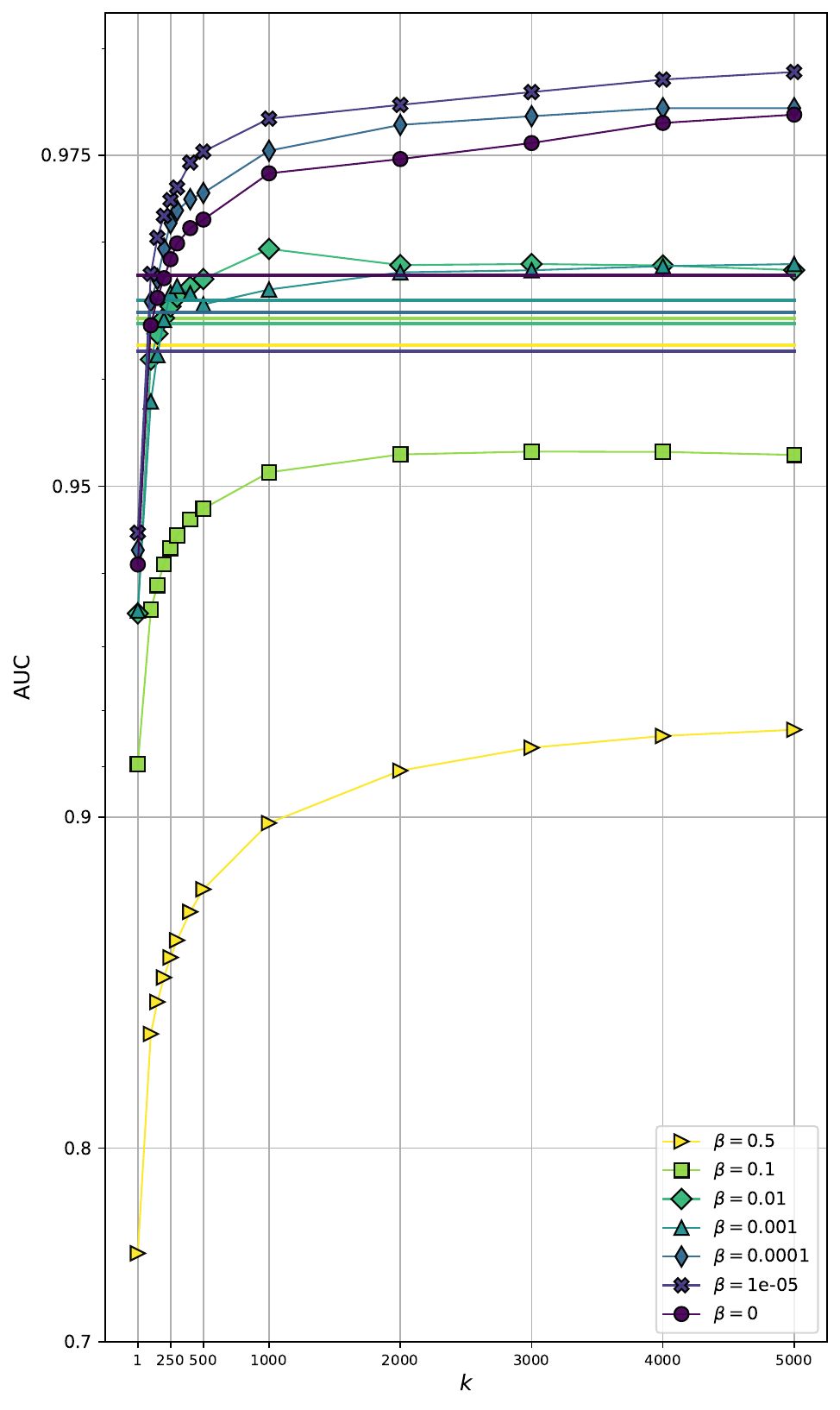}
    \caption{Mean AUROC of $\mathcal{L}_{rec}$-classification and $\mathcal{Z}_{k}$-classification with variables $\beta$ and $k$}
    \label{fig:recvslatent}
\end{figure}

Figure~\ref{fig:recvslatent} and Table~\ref{tab:recvslatent} show the mean AUROC across $\beta$ and $k$, averaged over four runs.
We have tested the $\mathcal{Z}_{k}$-classification with $k$ values of $1$, $100$, $150$, $200$, $250$, $300$, $400$, $500$, $1000$, $2000$, $3000$, $4000$, and $5000$.
The $\beta$ parameter was tested with values of $0$, $0.00001$, $0.0001$, $0.001$, $0.01$, $0.1$, and $0.5$.
In Table~\ref{tab:recvslatent}, bold values represent the best mean result per $\beta$.
Underlined values indicate cases where the mean AUROC with $\mathcal{Z}_{k}$-classification outperforms $\mathcal{L}_{rec}$-classification for a specific $\beta$.

In general, increasing $k$ improves the AUROC for the $\mathcal{Z}_{k}$-classification method.
Results show that $\mathcal{Z}_{k}$-classification can outperform $\mathcal{L}_{rec}$-classification in some cases with large value of $k$.
For $\mathcal{L}_{rec}$-classification, the best mean AUROC is achieved with $\beta=0$, and this method appears relatively insensitive to $\beta$; AUROC ranges from $0.962$ to $0.968$.
With this $\beta$ setting, $\mathcal{L}_{rec}$-classification is outperformed by $\mathcal{Z}_{k}$-classification for $k \ge 200$.
For $\mathcal{Z}_{k}$-classification, the best mean AUROC is obtained with $\beta=10^{-5}$ and $k=5000$.

\begin{table*}[h]
    \caption{Mean AUROC of $\mathcal{L}_{rec}$-classification and $\mathcal{Z}_{k}$-classification with variables $\beta$ and $k$}
    \label{tab:recvslatent}
    \begin{tabular*}{\linewidth}{@{\extracolsep{\fill}} rccccccccccccc|c @{}}
        
    \toprule
    \multirow[c]{2}[2]{*}{$\beta$} & \multicolumn{14}{c}{AUROC (\%)} \\
    \cmidrule(lr){2-15} 
    & $\mathcal{Z}_{1}$ & $\mathcal{Z}_{100}$ & $\mathcal{Z}_{150}$ & $\mathcal{Z}_{200}$ & $\mathcal{Z}_{250}$ & $\mathcal{Z}_{300}$ & $\mathcal{Z}_{400}$ & $\mathcal{Z}_{500}$ & $\mathcal{Z}_{1000}$ & $\mathcal{Z}_{2000}$ & $\mathcal{Z}_{3000}$ & $\mathcal{Z}_{4000}$ & $\mathcal{Z}_{5000}$ & $\mathcal{L}_{rec}$ \\ 
    \midrule
    0 & 94.11 & 96.43 & 96.63 & 96.76 & \underline{96.89} & \underline{96.99} & \underline{97.09} & \underline{97.14} & \underline{97.40} & \underline{97.48} & \underline{97.56} & \underline{97.66} & \textbf{\underline{97.70}} & 96.78 \\
    0.00001 & 94.49 & \underline{96.79} & \underline{97.03} & \underline{97.16} & \underline{97.25} & \underline{97.32} & \underline{97.46} & \underline{97.52} & \underline{97.68} & \underline{97.75} & \underline{97.81} & \underline{97.87} & \textbf{\underline{97.90}} & 96.23 \\
    0.0001 & 94.28 & \underline{96.60} & \underline{96.76} & \underline{96.96} & \underline{97.12} & \underline{97.19} & \underline{97.26} & \underline{97.29} & \underline{97.52} & \underline{97.65} & \underline{97.70} & \textbf{\underline{97.73}} & \underline{97.73} & 96.52 \\
    0.001 & 93.51 & 95.81 & 96.20 & 96.47 & \underline{96.66} & \underline{96.71} & \underline{96.66} & 96.58 & \underline{96.69} & \underline{96.80} & \underline{96.82} & \underline{96.85} & \textbf{\underline{96.86}} & 96.61 \\
    0.01 & 93.47 & 96.16 & 96.37 & \underline{96.48} & \underline{96.57} & \underline{96.64} & \underline{96.71} & \underline{96.76} & \textbf{\underline{96.96}} & \underline{96.85} & \underline{96.86} & \underline{96.85} & \underline{96.82} & 96.44 \\
    0.1 & 91.05 & 93.52 & 93.85 & 94.11 & 94.31 & 94.46 & 94.64 & 94.76 & 95.14 & 95.32 & 95.35 & 95.35 & 95.32 & \textbf{96.48} \\
    0.5 & 75.08 & 84.26 & 85.28 & 86.01 & 86.59 & 87.06 & 87.81 & 88.37 & 89.88 & 90.93 & 91.35 & 91.56 & 91.67 & \textbf{96.28} \\
    
    \bottomrule
    \end{tabular*}
 \end{table*}

\begin{figure}[h] \centering
    \includegraphics[width=6.6cm]{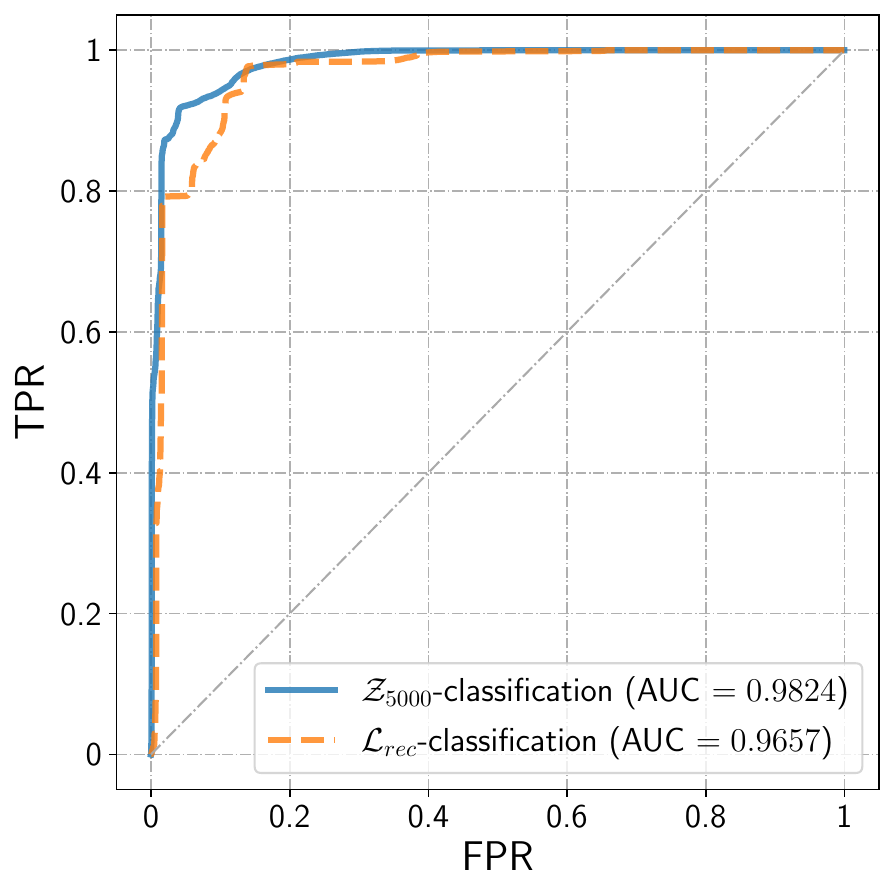}
    \caption{ROC curves for the binary classification task with $\mathcal{L}_{rec}$-classification and $\mathcal{Z}_k$-classification}
    \label{fig:global_roc_curve}
\end{figure}

\begin{figure}[h] \centering
    \includegraphics[height=6.6cm]{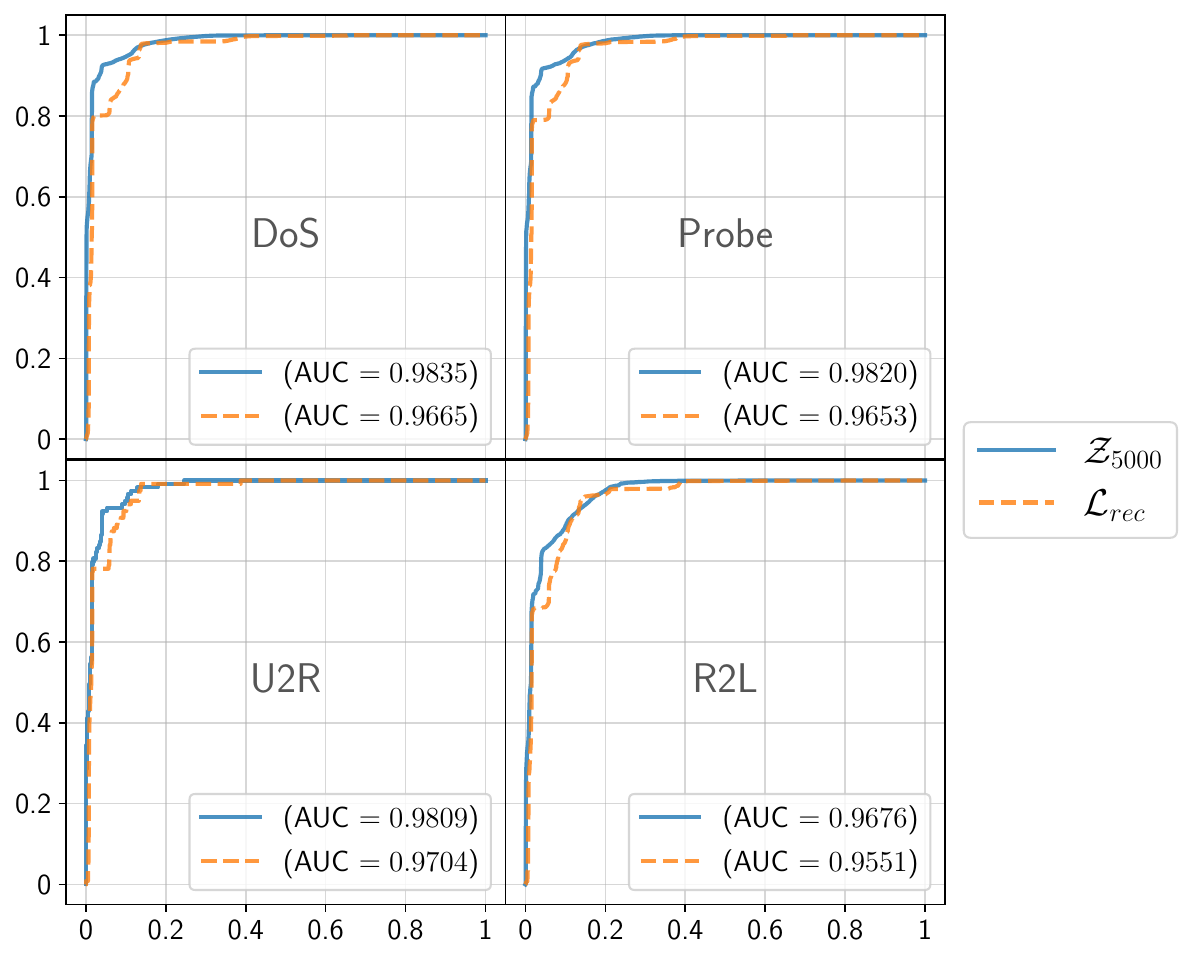}
    \caption{ROC curves on $X_{test}$ and $X_{attack}$ with $\mathcal{L}_{rec}$-classification and $\mathcal{Z}_{5000}$-classification, per attack class}
    \label{fig:roc_curve_per_class}
\end{figure}

\begin{figure}[h] \centering
    \includegraphics[width=8.5cm]{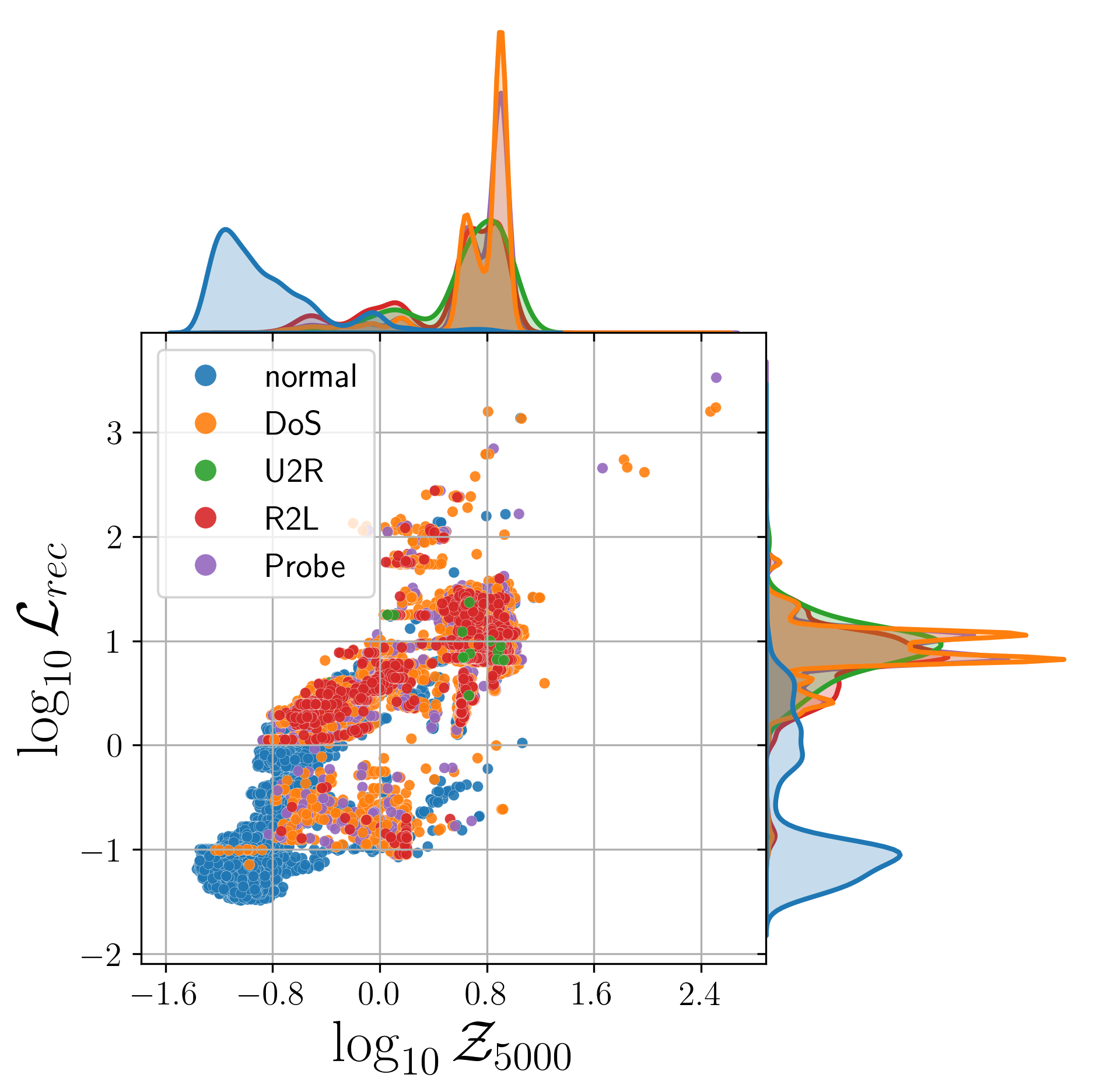}
    \caption{
        Distribution of $\mathcal{Z}_{5000}$-classification and $\mathcal{L}_{rec}$-classification on $X_{test}$ and $X_{attack}$.
        Blue points are classified as normal, purple as \textbf{Probe}, orange as \textbf{DoS}, green as \textbf{U2R}, and red as \textbf{R2L}.
        The distribution of each category is represented as a density on the opposing axes.
    }
    \label{fig:jointplot_distance_reconstruction}
\end{figure}

For the rest of the result we choose to focus on the results of a model trained with $\beta = 10^{-5}$ and $k = 5000$. The performance of this model is highlighted in Figure~\ref{fig:global_roc_curve}, Figure~\ref{fig:roc_curve_per_class} as ROC curves anlysis and in Figure~\ref{fig:jointplot_distance_reconstruction} to show the distribution of data classified by both methods.

Figure~\ref{fig:global_roc_curve} shows the ROC curves for the two methods on the binary classification task.

Figure~\ref{fig:roc_curve_per_class} shows the ROC curves for the two anomaly detection methods per attack class (Probe, DoS, U2R, and R2L) described in Section~\ref{sec:nsl_kdd}.
Some attack classes are more difficult to detect than others.

Figure~\ref{fig:jointplot_distance_reconstruction} shows the distribution of data classified by both methods.
Both approaches achieve excellent results.
The normal distribution is clearly separated from the attack distribution in both methods.
We can also see that sometimes certain normal data are well classified by one method and not the other.
This suggests the two methods are not redundant and can be complementary.
In fact, it is possible to implement an adaptive thresholding mechanism that considers both methods to classify the data.

\section{Conclusion and perspectives} \label{sec:conclusion}

This work studied unsupervised anomaly detection on NSL-KDD with a $\beta$-VAE, by comparing two decision signals based on different principles: reconstruction-based scoring ($\mathcal{L}_{rec}$-classification) and latent space distance-based scoring computed as the mean Euclidean distance to the $k$ nearest neighbors ($\mathcal{Z}_{k}$-classification). We showed that latent distance can match or surpass reconstruction error depending on $\beta$ and $k$, while the two signals remain complementary for some samples.

The latent-space method enables incremental learning. Because decisions rely on reference embeddings, the model can be updated online by appending new normal and labeled abnormal projections without retraining the $\beta$-VAE. This makes it possible to adapt to evolving operating conditions and to progress from anomaly detection to behavior classification: clusters of latent patterns corresponding to distinct operating modes can be tracked and labeled over time, enabling fine-grained classification of behaviors in addition to binary anomaly flags.

Beyond Euclidean distance, we can also consider a Mahalanobis score in latent space. Rather than sampling $z$ from $q_\phi(z|x)$, we simply use the encoder outputs as a deterministic embedding, the mean $\mu(x)$ and evaluate a Mahalanobis distance to the normal reference statistics estimated on $Z_{train}$ \cite{mclachlan1999mahalanobis}. The advantage of Mahalanobis distance over Euclidean distance is that it takes into account the covariance structure of the data, which can be particularly useful in high-dimensional spaces where features may be correlated, as discussed in \cite{pitsiorlasTrustworthy2024}.

Future work will focus on fusing reconstruction-based and latent-based scores via calibrated or learned aggregation; implementing and benchmarking Mahalanobis-based detectors in the latent space; broadening the evaluation to diverse datasets and operating conditions; and leveraging incremental learning to track latent clusters and enable behavior-aware intrusion analysis.

\bibliographystyle{IEEEtran}
\bibliography{main}

\begin{thebibliography}{10}
\providecommand{\url}[1]{#1}
\csname url@samestyle\endcsname
\providecommand{\newblock}{\relax}
\providecommand{\bibinfo}[2]{#2}
\providecommand{\BIBentrySTDinterwordspacing}{\spaceskip=0pt\relax}
\providecommand{\BIBentryALTinterwordstretchfactor}{4}
\providecommand{\BIBentryALTinterwordspacing}{\spaceskip=\fontdimen2\font plus
\BIBentryALTinterwordstretchfactor\fontdimen3\font minus
  \fontdimen4\font\relax}
\providecommand{\BIBforeignlanguage}[2]{{%
\expandafter\ifx\csname l@#1\endcsname\relax
\typeout{** WARNING: IEEEtran.bst: No hyphenation pattern has been}%
\typeout{** loaded for the language `#1'. Using the pattern for}%
\typeout{** the default language instead.}%
\else
\language=\csname l@#1\endcsname
\fi
#2}}
\providecommand{\BIBdecl}{\relax}
\BIBdecl

\bibitem{tavallaee2009detailed}
M.~Tavallaee, E.~Bagheri, W.~Lu, and A.~A. Ghorbani, ``A detailed analysis of
  the {{KDD CUP}} 99 data set,'' in \emph{2009 {{IEEE Symposium}} on
  {{Computational Intelligence}} for {{Security}} and {{Defense
  Applications}}}, Jul. 2009, pp. 1--6.

\bibitem{higgins2017beta}
I.~Higgins, L.~Matthey, A.~Pal, C.~Burgess, X.~Glorot, M.~Botvinick,
  S.~Mohamed, and A.~Lerchner, ``beta-vae: Learning basic visual concepts with
  a constrained variational framework,'' in \emph{International conference on
  learning representations}, 2017.

\bibitem{kullback1951information}
S.~Kullback and R.~A. Leibler, ``On information and sufficiency,'' \emph{The
  Annals of Mathematical Statistics}, vol.~22, no.~1, pp. 79--86, 1951.

\bibitem{kingma2022autoencoding}
D.~P. Kingma and M.~Welling, ``Auto-encoding variational bayes,'' 2022.

\bibitem{HeFault2007}
Q.~He and J.~Wang, ``Fault detection using the k-nearest neighbor rule for
  semiconductor manufacturing processes,'' \emph{IEEE Transactions on
  Semiconductor Manufacturing}, vol.~20, no.~4, pp. 345--354, 2007.

\bibitem{SongHybrid2017}
H.~Song, Z.~Jiang, A.~Men, and B.~Yang, ``A {{Hybrid Semi-Supervised Anomaly
  Detection Model}} for {{High-Dimensional Data}},'' \emph{Computational
  Intelligence and Neuroscience}, vol. 2017, no.~1, p. 8501683, 2017.

\bibitem{guoAnomaly2018}
J.~Guo, G.~Liu, Y.~Zuo, and J.~Wu, ``An {{Anomaly Detection Framework Based}}
  on {{Autoencoder}} and {{Nearest Neighbor}},'' in \emph{2018 15th
  {{International Conference}} on {{Service Systems}} and {{Service
  Management}} ({{ICSSSM}})}, Jul. 2018, pp. 1--6.

\bibitem{angiullilatent2023}
F.~Angiulli, F.~Fassetti, and L.~Ferragina, ``$\mathrm{Latent}{Out}$: An
  unsupervised deep anomaly detection approach exploiting latent space
  distribution,'' \emph{Machine Learning}, vol. 112, no.~11, pp. 4323--4349,
  Nov. 2023.

\bibitem{zhangAutomated2018}
Z.~Zhang, T.~Jiang, S.~Li, and Y.~Yang, ``Automated feature learning for
  nonlinear process monitoring -- {{An}} approach using stacked denoising
  autoencoder and k-nearest neighbor rule,'' \emph{Journal of Process Control},
  vol.~64, pp. 49--61, Apr. 2018.

\bibitem{corizzoAnomaly2019}
R.~Corizzo, M.~Ceci, and N.~Japkowicz, ``Anomaly {{Detection}} and {{Repair}}
  for {{Accurate Predictions}} in {{Geo-distributed Big Data}},'' \emph{Big
  Data Research}, vol.~16, pp. 18--35, Jul. 2019.

\bibitem{ramakrishnaEfficient2021}
S.~Ramakrishna, Z.~Rahiminasab, G.~Karsai, A.~Easwaran, and A.~Dubey,
  ``Efficient {{Out-of-Distribution Detection Using Latent Space}} of
  {$\beta$}-{{VAE}} for {{Cyber-Physical Systems}},'' \emph{ACM Trans.
  Cyber-Phys. Syst.}, vol.~6, no.~2, Apr. 2022.

\bibitem{pitsiorlasTrustworthy2024}
I.~Pitsiorlas, G.~Arvanitakis, and M.~Kountouris, ``Trustworthy {{Intrusion
  Detection}}: {{Confidence Estimation Using Latent Space}},'' \emph{2024 22nd
  International Symposium on Modeling and Optimization in Mobile, Ad Hoc, and
  Wireless Networks (WiOpt)}, pp. 92--98, 2024.

\bibitem{astridConstricting2024}
M.~Astrid, M.~Z. Zaheer, and S.~Lee, ``Constricting {{Normal Latent Space}} for
  {{Anomaly Detection}} with {{Normal-only Training Data}},'' in \emph{5th
  Workshop on practical ML for limited/low resource settings}, 2024.

\bibitem{kamnitsasSemi2018}
K.~Kamnitsas, D.~Castro, L.~L. Folgoc, I.~Walker, R.~Tanno, D.~Rueckert,
  B.~Glocker, A.~Criminisi, and A.~Nori, ``Semi-{{Supervised Learning}} via
  {{Compact Latent Space Clustering}},'' in \emph{Proceedings of the 35th
  International Conference on Machine Learning}, ser. Proceedings of Machine
  Learning Research, J.~Dy and A.~Krause, Eds., vol.~80.\hskip 1em plus 0.5em
  minus 0.4em\relax PMLR, 10--15 Jul 2018, pp. 2459--2468.

\bibitem{kingmaAdam2017}
\BIBentryALTinterwordspacing
D.~P. Kingma and J.~Ba, ``Adam: {A} method for stochastic optimization,'' in
  \emph{3rd International Conference on Learning Representations, {ICLR} 2015,
  San Diego, CA, USA, May 7-9, 2015, Conference Track Proceedings}, Y.~Bengio
  and Y.~LeCun, Eds., 2015. [Online]. Available:
  \url{http://arxiv.org/abs/1412.6980}
\BIBentrySTDinterwordspacing

\bibitem{mclachlan1999mahalanobis}
G.~J. McLachlan, ``Mahalanobis distance,'' \emph{Resonance}, vol.~4, no.~6, pp.
  20--26, 1999.

\end{thebibliography}

\end{document}